# A Novel Low-cost FPGA-based Real-time Object Tracking System

Peng Gao[1], Ruyue Yuan[1], Zhicong Lin[1], Linsheng Zhang[2], Yan Zhang[1]

[1] Shenzhen Graduate School, Harbin Institute of Technology, China
[2] Sanechips Technology Co., Ltd, China

**Abstract**

In current visual object tracking system, the CPU or GPU-based visual object tracking systems have high computational cost and consume a prohibitive amount of power. Therefore, in this paper, to reduce computational burden of Camshift algorithm, we propose a novel visual object tracking algorithm by exploiting the properties of binary classifier and Kalman predictor. Moreover, we present a low-cost FPGA-based real-time object tracking hardware architecture. Extensive evaluations on OTB benchmark demonstrate that the proposed system has extremely compelling real-time, stability and robustness. The evaluation results show that the accuracy of our algorithm is about 48%, and the average speed is about 309 frames per second.

## 1. Introduction

Object tracking is a fundamental task in computer vision, with applications in video surveillance, human-machine interfaces and robot perception. In general, object tracking requires high accuracy and involves a lot of transcendental functions and high-precision floating-point operations, so the target tracking technology used in the industry is mostly based on CPU or GPU software programming. Nevertheless, these implementations have high computational cost and consume a prohibitive amount of energy. As a kind of hardware platform, FPGA has advantages of fast speed and strong stability for digital image processing. It has significant application value to use FPGA as the hardware platform to develop moving object tracking system.

The developments on object visual tracking hardware systems are encouraged continuously over an extended period. For example, Liu et al. [2] incorporate a background subtraction-based algorithm to implement an object tracking system, which can reach more than 100X for complex video inputs. Additionally, Elkhatib et al. [3] propose an object tracking hardware system which can deal with the 640x480 input images run on a 20MHz processor. Furthermore, Singh et al. [4] concentrate on the integration of modified particle filters and SAD for object tracking, their implemented system can robustly track the objects based on Xilinx Virtex V. All of the hardware systems mentioned above have achieved significant tracking results. Obviously, the more complex algorithm, the better tracking efficiency of a hardware system.

In this paper, we propose a novel low-cost real-time object tracking system based on FPGA. In order to solve numerous floating-point calculations involved in the reverse projection process in Camshift algorithm [5], an optimization method is also proposed. Initially, a binary classifier and a Kalman predictor are used to achieve the fast separation and dual-mode tracking of the region of interesting (ROI) respectively. Then we implement the hardware platform and give the hardware and software experiments and comparison results. The system workflow as shown in Figure 1.

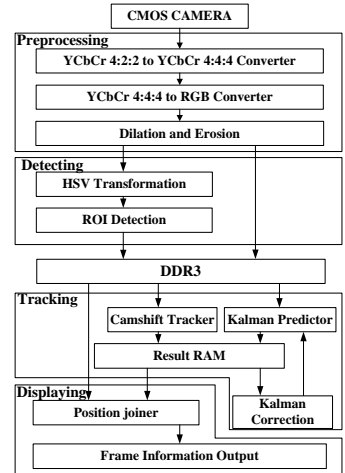

Figure 1. Sytem Workflow

The paper is organized into 5 sections. Section 2 introduces a novel tracking algorithm based on the classical Camshift algorithm. Section 3 presents the implementation details of our hardware object tracking system. Section 4 evaluates our hardware system extensively. Section 5 gives some concluding remarks and directions for future work.

## 2. Algorithm and Theory

In this paper, the object tracking algorithm is optimized based on the Camshift algorithm [5]. The binary classifier and the Kalman predictor are employed to accelerate separation and dual-mode tracking of the ROI. The algorithm is optimized for the hardware implementation. The algorithm includes video image format converting, color space converting and visual tracking.

**2.1 Getting ROI Value**

In classical Camshift algorithm, the threshold masking and the raw image mapping operations both suffer from a high computational complexity by the process of selecting ROI. It is not suitable for low-cost hardware implementation. Considering the real-time and hardware implementation of the system, a novel visual object tracking algorithm using the binary classifier to obtain ROI from the HSV space without the reverse projection is proposed in this paper. The data width of each pixel is reduced from 24 bits to 1 bit, both the bulk of data and the cost of implementation are decreased. Here the thresholds are set according to the mean value by the HSV of the tracking object.

In terms of the selected characteristics of the initial bounding box, the binary classifier separates the ROI from the image defined in Eq.1. In subsequent operations, we use the classified binary image reduce the computational cost efficiently.

$$\begin{cases} H_{kc} = 1, & if \quad |H_k - \overline{H}_{k-1}| < H_T \\ S_{kc} = 1, & if \quad |S_k - \overline{S}_{k-1}| < S_T \\ V_{kc} = 1, & if \quad |V_k - \overline{V}_{k-1}| < V_T \end{cases} \quad (1)$$

where $\overline{H}_{k-1}, \overline{S}_{k-1}, \overline{V}_{k-1}$ are the HSV component mean of the pixels in the previous frame bounding box respectively. $H_T$, $S_T$, $V_T$ are preset thresholds. $H_{kc}$, $S_{kc}$, $V_{kc}$ are the classification value of the current image pixel. Set them to 1 if they belong to the positive samples of ROI.

To prevent threshold drift, the HSV weighted classifier is proposed as Eq.2.

$$A_{kc} = 1, \quad if \quad W(\text{HSV}) < A_T \quad (2)$$

where $W(\text{HSV}) = (\alpha|H_k - \overline{H}_{k-1}| + \beta|S_k - \overline{S}_{k-1}| + \gamma|V_k - \overline{V}_{k-1}|)$ is the weight formula for HSV component. $\alpha$, $\beta$, $\gamma$ are the weight coefficient of each component ($\alpha + \beta + \gamma = 1$). The default image trigger signal is static, only if $H_{kc}$, $S_{kc}$, $V_{kc}$, $A_{kc}$ are set to 1 at same time, the current pixel is classified as ROI, that is

$$ROI_{kc} = H_{kc} \& S_{kc} \& V_{kc} \& A_{kc} \quad (3)$$

**2.2 Tracking Algorithm**

In the standard Camshift tracking, the traditional mean-shift algorithm uses the one-moment and zero-moment to calculate center position of the prediction bounding box. However, when there is background clutters or partial occlusion, the confounding noise will be calculated and will make the tracking drift. To solve this problem, we propose a method of weighting the moment of calculation, as shown in Eq.4.

$$\begin{cases} M_{10} = \sum_x W(I(x, y)) \\ M_{01} = \sum_y W(I(x, y)) \\ M_{00} = \sum_x \sum_y W(I(x, y)) \end{cases} \quad (4)$$

where $W(\cdot)$ represents a weighted function, and $I(\cdot)$ denotes the classification value of the pixel. The closer the current pixel is to the center of the previous frame, the larger the weight.

Camshift algorithm has an extremely compelling effect when tracking object which moves in small range. But it becomes less effective when object moving in large range, as opposed to Kalman predictor. Therefore, based on the characteristics of the two algorithms, we optimized the classical Camshift algorithm by combined binary classifier with Kalman predictor. When using the Camshift algorithm for objects detecting, if the size of the current bounding box is compared with the size of the bounding box in the previous frame too large or small, the Kalman predictor is used to correct the prediction boundary box.

In traditional software implementations, to calculate the centroid of the bounding box, we need a serial iterative execution of the meanShift algorithm until the results converge or reach the threshold of the iteration. But it is a very expensive approach and cannot be operated in real-time. Tracking high-speed moving objects are prone to drop frames. In order to eliminate the long-time error which may cause the object drift [1], a parallel Camshift algorithm is proposed, parallelize the convergence iteration and calculate all the proposal objects position $x_i$ simultaneously. When the input image size is $M \times N$, the time complexity of the linear algorithm is reduced from $O(MN)$ to $O(\log_n MN)$, where $n$ is the parallel threads number.

**3. Hardware Implementation**

In this paper, we implement our hardware system based on Xilinx Spartan-6 platform.

**3.1 System Architecture**

System hardware configuration is shown in Table I.

Table I. System hardware configuration

| Hardware | Configuration |
|---|---|
| FPGA | Xilinx Spartan-6 |
| System clock | 148.5MHz |
| DDR3 | 2Gbit |
| Maximum bandwidth | 10 Gbit/s |

The raw image resolution and captured speed are set to 1080P and 60FPS respectively.

### 3.2 Implementation Detail

In order to save the bandwidth of FPGA, the video image format and resolution are set to YCbCr 4:2:2 and 1080P, respectively. Firstly, we convert the camera data into the RGB format. Next, we process the video by using Channel 0. Resort to ping-pong operation, we can deal with four-frame images simultaneously. Finally, we use the HDMI interface to display the bounding box. The RGB images are converted to the HSV color space, then we use Channel 1 to write the ROI data which are determined by binary classifier into DDR3. After transferring the ROI binary image data to the object tracking algorithm circuit module from DDR3, the algorithm circuit module calculates the pixel position of the top-left and the bottom-right corner of the bounding box respectively and transfers the results to the bounding box circuit module. Then, Channel 2 of DDR3 is used to combine the tracking window to the corresponding position. During the system design process, the read operation of the 3 channels above of DDR3 is executed in parallel. When the n $^{th}$ frame is converted and marked, the tracking of the (N-1) $^{th}$ frame is performed and displaying the (N-2) $^{th}$ frame.

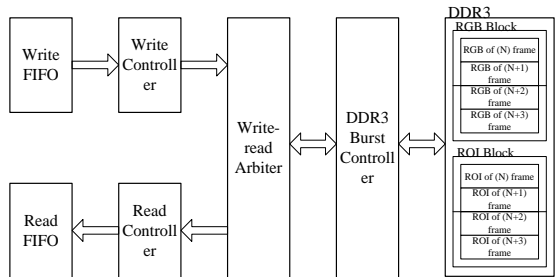

Figure 2. Optimization of DDR.

The search stage uses the 1bit ROI data which outputted from preprocessing stage. In order to meet the requirements of the data throughput, the usage of the DDR3 is as shown in Figure 2, it shows the DDR3 is divided into 2 regions, each region can store 4 frames of 1080P images. During the operation process, we group 2 frame images, thus each group not only can be addressing by the ping-pong operation at the same time, but also meet the requirements of the system parallel operation and improve system performance. As the camera output video image data is serial continuous input, the system uses a large number of floating-point calculation. In order to avoid either the occurrence of the non-real-time or the drop frame phenomenon when using the software platforms, a multistage pipeline is used for format conversion and ROI classification during the pre-process and detection processing. Although the system can track a group of images simultaneously, considering that most of the high-speed cameras' shutter time are 1/60s, and generally no more than 1/120s, only one frame can be tracked at a time to meet low-cost requirements for the system.

When calculating the Camshift algorithm, we use the parallel algorithm mentioned in Section 2.3. The result is storing to the Result RAM 1, as shown in Figure 3.

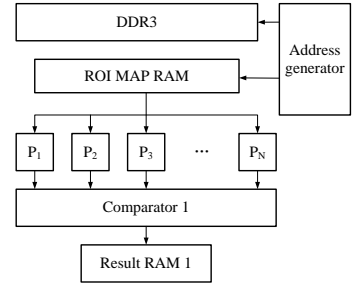

Figure 3. Parallel operating of algorithm module.

The Kalman predictor is used to estimate the object motion state and the results are writing into the Result RAM 2. Then compare the current bounding box's size of Camshift calculating with the previous. If the comparison result is greater than the preset threshold, the Kalman predictor is used to correct the results, otherwise the Camshift calculation result is used as the output. The final output is the top-left and bottom-right pixel positions of the bounding box.

### 4. Results

In order to verify the accuracy of our real-time visual object tracking algorithm, we use Matlab to do verification. Figure 4 shows the experimental results for the *girl* in the literature [1] and our HD image sequences. At present, the classifier's threshold used by our algorithm is set for a single object feature and it cannot be applied to multi-objects tracking. The following results are valid for a single object in the general background

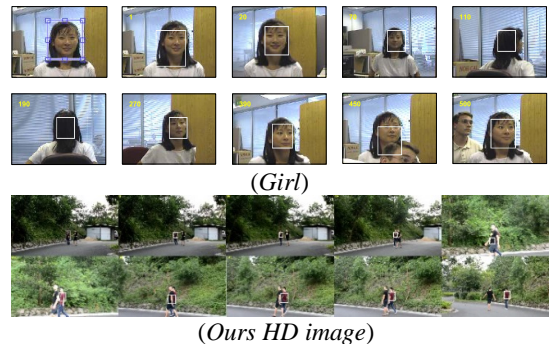

(*Girl*)

(*Ours HD image*)
Figure 4. Our algorithm tracking evaluations

Figure 5 shows the overlap rate and accurate evaluation of our algorithm and several state-of-the-art trackers mentioned by Wu et al. [1] which performs significantly, All the experiments are performed All experiments are conducted on an Intel i5-4590 CPU at 3.3GHz with 8GB RAM on the segmental sequences of OTB50 and our HD video sequences. The average speed of our algorithm and eight state-of-the-art trackers are shown in Table II.

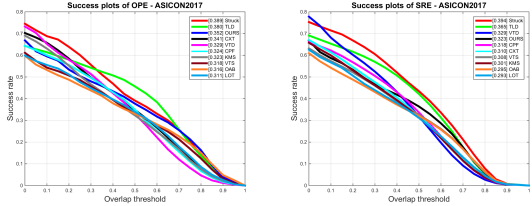

Figure 5. The success plots of our tracker and other state-of-the-art trackers

Table II. Tracking speed of our tracker and other state-of-the-art trackers

| Tracker | Precision | Mean FPS |
|---|---|---|
| Struck | 53.5% | 9.8 |
| TLD | 51.9% | 24.4 |
| OURS | 48.4% | **309.91** |
| CXT | 45.4% | 13.9 |
| VTD | 44.4% | - |
| CPF | 43.1% | - |
| VTS | 42.2% | 6.3 |
| LOT | 40.7% | 0.5 |
| OAB | 40.5% | 5.5 |
| KMS | 38.9% | - |

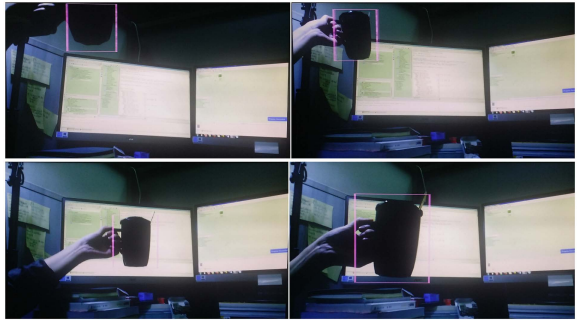

Figure 6. Hardware system experiment

Figure 6 shows the results of hardware system experiment. The test object is the black cup in the video image, through the figure we can see the system tracking process is obvious, and the tracking effect is still satisfactory when there is occlusion or interference. The system is capable of handling the moving object and the moving camera.
From Figure 4 to Figure 6, it concludes the accuracy of our algorithm is about 48%, and the mean speed is about 309 FPS. In the premise of ensuring a desirable accuracy, the calculation speed has been improved with more stability and robustness and ensured the optimization of hardware implement. From Table III we can get the FPGA resource utilization.

Table III. FPGA Resource Utilization Summary

| Resource | Used Resources | Utilization |
|---|---|---|
| Slice Registers | 16019 | 29% |
| Slice LUTs | 14630 | 53% |
| DSP48A1s | 42 | 72% |

## 5. Summary

In this paper, we have explored how to improve efficiency of object visual tracking on the basis of the classical Camshift algorithm. Evaluations on several benchmarks also demonstrate our tracker is more compact and much faster than several state-of-the-art trackers. The most considerable is that our algorithm can work in real-time and suitable for low-cost hardware implementations. In the future work, we plan to improve the recognition effect of the tracking system, so that the system can track in background clutters with more robustness and stability.

## Acknowledgments

This work is supported by the Science and Technology Planning Project of Guangdong Province (Grant No.2016B090918047).